\documentclass[10pt,twocolumn,letterpaper]{article}
\usepackage{bm}
\usepackage{float}

\usepackage{booktabs} 
\usepackage{authblk}

\usepackage[pagenumbers]{cvpr} 

\usepackage[dvipsnames]{xcolor}

\definecolor{cvprblue}{rgb}{0.21,0.49,0.74}
\usepackage[pagebackref,breaklinks,colorlinks,citecolor=cvprblue]{hyperref}

\begin{document}
\date{}

\title{DivAvatar: Diverse 3D Avatar Generation with a Single Prompt}

\author[1]{Weijing Tao}
\author[2]{Biwen Lei}
\author[1]{Kunhao Liu}
\author[1]{Shijian Lu}
\author[2]{Miaomiao Cui}
\author[2]{Xuansong Xie}
\author[1]{Chunyan Miao}

\affil[1]{Nanyang Technological University} \affil[2]{Alibaba Group}


\affil[ ]{\small\texttt{\{weijing002,kunhao001\}@e.ntu.edu.sg, \{shijian.lu, ascymiao\}@ntu.edu.sg}}
\affil[ ]{\texttt{\{biwen.lbw, miaomiao.cmm\}@alibaba-inc.com,xingtong.xxs@taobao.com }}



\twocolumn[{
    \renewcommand\twocolumn[1][]{#1}
    \maketitle
    \begin{center}
        \centering
        \captionsetup{type=figure}
        \includegraphics[width=0.9\textwidth]{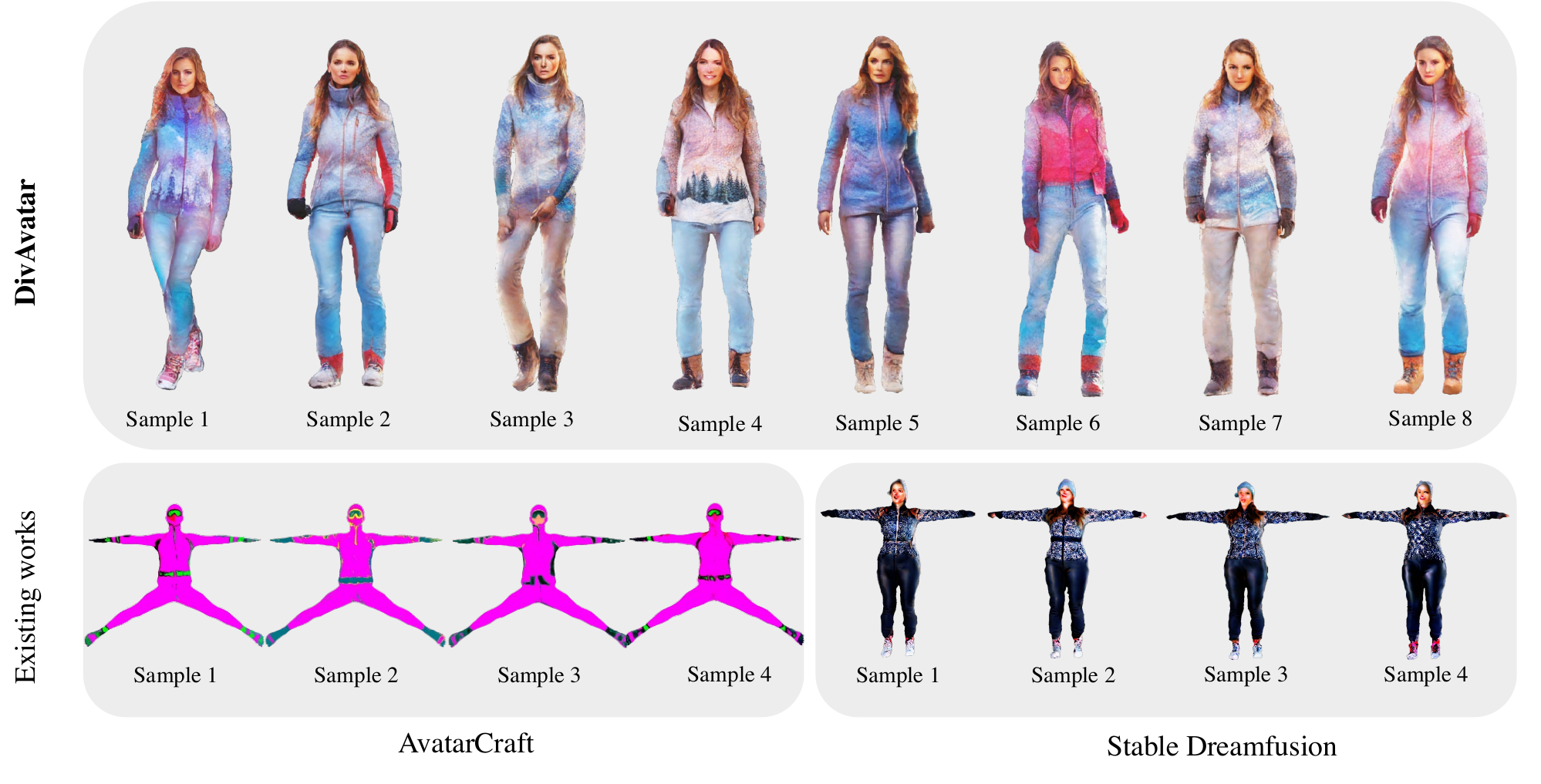}
        \captionof{figure}{Diverse generation of DivAvatar. \textit{Input text prompt: A woman wearing ski clothes.} Given a single text prompt, DivAvatar can generate a varied set of appearances, while existing works generates appearances with subtle differences. Furthermore, DivAvatar readily generates avatars in different sampled poses, while most existing works can only generate pose according to input pose and require additional articulation for more varied poses.
        }
        \label{fig:teaser}
    \end{center}
}]

\begin{abstract}

Text-to-Avatar generation has recently made significant strides due to advancements in diffusion models. However, most existing work remains constrained by limited diversity, producing avatars with subtle differences in appearance for a given text prompt. We design DivAvatar, a novel framework that generates diverse avatars, empowering 3D creatives with a multitude of distinct and richly varied 3D avatars from a single text prompt. Different from most existing work that exploits scene-specific 3D representations such as NeRF, DivAvatar finetunes a 3D generative model (i.e., EVA3D), allowing diverse avatar generation from simply noise sampling in inference time. DivAvatar has two key designs that help achieve generation diversity and visual quality. The first is a noise sampling technique during training phase which is critical in generating diverse appearances. The second is a semantic-aware zoom mechanism and a novel depth loss, the former producing appearances of high textual fidelity by separate fine-tuning of specific body parts and the latter improving geometry quality greatly by smoothing the generated mesh in the features space. Extensive experiments show that DivAvatar is highly versatile in generating avatars of diverse appearances.

\end{abstract}


\section{Introduction}
\label{sec:intro}

Automated generation of high-quality 3D avatars has gained significant interest thanks to its applications in video games, augmented and virtual reality, virtual try-ons, etc. With the recent advancement in Large Language Models \cite{raffel2020exploring, brown2020language} and Diffusion Models \cite{ho2020denoising, song2020denoising}, automating the generation of high-quality 3D avatars using only natural language descriptions holds great potential for saving time and labour resources. The integration of diversity is critical in the automated generation of avatars, playing a significant role in enhancing inclusivity within virtual environments and allowing representation of a broad spectrum of identities. Moreover, generating diverse avatars from a single text description not only streamlines the creative process but also markedly decreases production times.

 Recent text-to-avatar studies \cite{zhang2023avatarverse,kolotouros2023dreamhuman,cao2023dreamavatar,jiang2023avatarcraft} exploit text-guided diffusion models and Score Distillation Sampling (SDS) \cite{poole2022dreamfusion} to optimize Neural Radiance Fields (NeRFs) \cite{mildenhall2021nerf}. They capitalize on human body modelling \cite{loper2023smpl} for accurate 3D human geometry, but often neglect the equally important aspect of appearances. Specifically, these works adopt SDS loss as a primary means to dictate avatar appearances, which by nature tends to converge to less varied outcomes \cite{poole2022dreamfusion}. As a result, the generation of avatars using current techniques fall short in capturing the full diversity of human identities, presenting challenges for 3D creators striving for realistic diversity in virtual environments. Beyond that, the existing works involve training of scene-specific NeRF, which limits their scalability. They typically generate one avatar at a time \cite{zhang2023avatarverse,kolotouros2023dreamhuman,cao2023dreamavatar,jiang2023avatarcraft,wang2023prolificdreamer}, which is inefficient for industries needing quick generation of a breadth of unique characters that fulfil a given text prompt quickly.

We propose DivAvatar, a novel avatar generation framework that enables diverse appearances from one text prompt. DivAvatar incorporates diffusion prior and finetunes a pretrained EVA3D \cite{hong2022eva3d}, a compositional human NeRF representation constructed as a GAN framework. Besides the geometry prior on human modeling \cite{loper2023smpl}, EVA3D also serves as a strong appearance prior from its training data including a wide array of human images. Hence, DivAvatar exploits the intrinsic appearance prior of EVA3D, providing the foundation for generating avatars with lifelike and diverse appearances. Additionally, we finetune a generative model instead of a scene-specific model like NeRF. This allows us to effectively generate diverse avatars at inference time via noise sampling, where each noise generates a unique avatar appearance that aligns with the text input. This greatly reduces the time taken for diverse generation as compared to existing work which would have to repeat the full training cycle. 

DivAvatar consists of two novel designs for achieving generative diversity and generation quality. The first is a noise sampling technique that effectively resurrects the GAN's capability for diverse avatar generation despite the nature of SDS. This strategic adjustment proficiently generates a spectrum of avatar visualizations corresponding to a singular textual descriptor via noise sampling during the inference time. The second is a comprehensive semantic aware zoom mechanism that aims to ensure textual fidelity which tends to be compromised when the avatar's appearance does not accurately reflect the descriptive text. In addition, we design a novel depth loss that effectively refines the geometry quality in the feature space. Overall, DivAvatar leverages a more comprehensive set of priors than existing avatar works, and enables efficient and rapid generation of a wide array of avatars. The spectrum of avatars generated not only exhibit a high degree of diversity but also maintain textual coherence with the original single training prompt.

The contributions of this work can be summarized in three aspects. \textit{First}, we propose a novel pipeline that can create diverse 3D human avatars rapidly that can be placed in a variety of poses from a single input text prompt. \textit{Second}, we design a noise sampling technique that ensures generation diversity by generating a varied set of avatars that align with a single input text prompt at inference time. \textit{Third}, we design a strategic semantic zoom that improves adherence to complex text prompts, as well as a novel depth loss that enhances the quality of the generated geometry effectively in the feature space.


\section{Related Work}
\label{sec:related}

\subsection{Text-to-3D Generation}
The advancement in text-guided 2D image generation has set the foundation for 3D content creation based on textual descriptions. Pioneering works include CLIP forge \cite{sanghi2022clip}, DreamFields \cite{jain2022zero}, and CLIP-Mesh \cite{mohammad2022clip}, which utilize the renowned CLIP \cite{radford2021learning} to optimize underlying 3D representations, such as NeRF and textured meshes. DreamFusion \cite{poole2022dreamfusion} builds on top of DreamFields and proposes to use the SDS loss, derived from a pre-trained diffusion model \cite{saharia2022photorealistic} as supervision during optimization. Subsequent improvements over DreamFusion include optimizing 3D representations in a latent space \cite{metzer2023latent} and coarse-to-fine manner \cite{lin2023magic3d}. The closest work to us is ProlificDreamer \cite{wang2023prolificdreamer} , which introduces Variational Score Distillation to produce diverse results in general 3D objects. However, in the specific domain of avatar generation, these innovative methods still grapple with challenges like inferior quality, presence of the Janus problem, and flawed representation of anatomical structures. Furthermore, the maximum number of different appearances ProlificDreamer can generate at a time is up to four. On the other hand, our work is able to generate unlimited number of diverse appearances, through directly sampling noise from a normal distribution.

\subsection{Text-to-Avatar Generation}
To enable 3D avatar generation from text, several approaches have been proposed. Avatar-CLIP \cite{hong2022avatarclip} sets the foundation by initializing human geometry with a shape VAE and utilizing CLIP \cite{radford2021learning} to assist in geometry and texture generation. DreamAvatar \cite{cao2023dreamavatar} and AvatarCraft \cite{jiang2023avatarcraft} integrate the human parametric model with pre-trained 2D diffusion models for 3D avatar creation. DreamHuman \cite{kolotouros2023dreamhuman} further introduces a camera zoom-in technique to increase resolution of resulting avatars; while DreamWaltz \cite{huang2023dreamwaltz} incorporates a skeleton-conditioned ControlNet \cite{zhang2023adding} and develops an occlusion-aware SDS guidance for pose-aligned supervision. Despite achieving impressive results, these methods are limited by weak SDS guidance and scarce skeleton conditioning, hampering their ability to produce avatars with consistent multi-view appearances and pose control. Therefore a more recent study, AvatarVerse \cite{zhang2023avatarverse} and a concurrent work, \cite{masteravatarstudio} employ DensePose-conditioned ControlNet for SDS guidance for a more stable, view-consistent avatar creation and pose control. Unlike these approaches, our method gravitates towards a finetuned generative model with a strong 3D-aware prior, presenting an alternative approach for ensuring dependable avatar generation of any pose. Additionally, a recurring limitation across existing methodologies is that they generate avatars with limited variation from a single text prompt. In contrast, our method stands out by enabling the generation of distinctly diverse results through sampling from the fine-tuned generative model, further simplifying the process of text-driven avatar creation and expanding creative possibilities.

\section{Method}
\label{sec:method}

\begin{figure*}[ht!]
  \centering
  \includegraphics[width=0.85\linewidth]{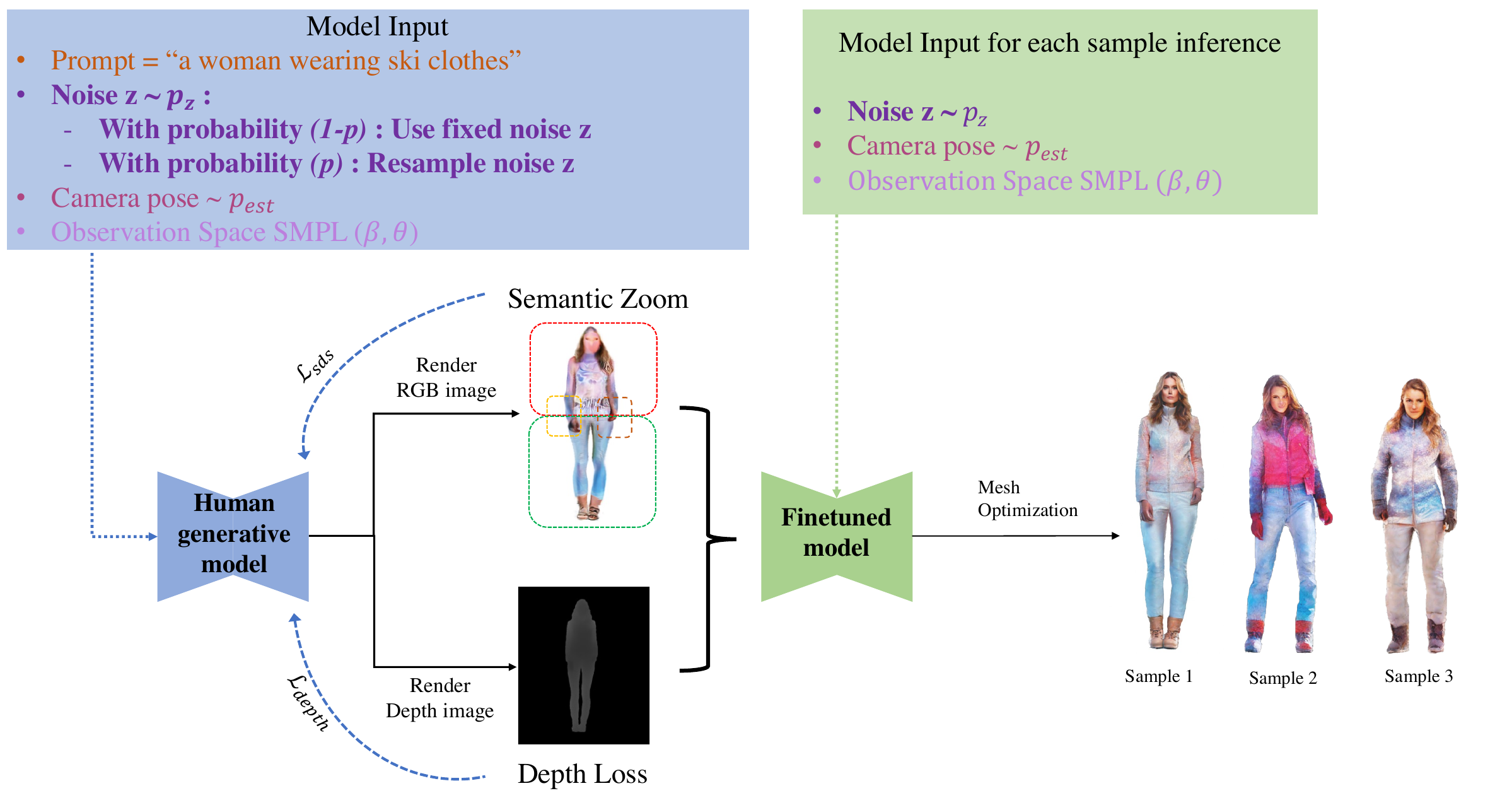}
  \caption{\textbf{Overview of DivAvatar.} We generate a set of diverse avatars that well align with the input text prompt using a strategic noise sampling in the process of finetuning the pre-trained human generative model (i.e. EVA3D). At each training iteration, the noise used as input is either randomly sampled or fixed. The optimisation of the avatar is guided by Score Distillation Sampling loss with our semantic aware zoom technique that ensures higher textual fidelty, and a feature-based depth loss that smooths the geometry. At inference time, DivAvatar is able to generate numerous avatars of different appearance in various pose by sampling the noise and SMPL parameters.}
  \label{fig:pipeline}
\end{figure*}

In this section, we present DivAvatar, a novel pipeline that generates a set of realistic 3D avatars of any pose from a single text description. After introducing some preliminaries, we first explain the training strategy enabling diverse avatar generation. We then introduce novel strategies that enhance the synthesis quality: the semantic zoom and feature based depth loss. Following these, we perform mesh optimization using DMTet finetune.

\subsection{Preliminaries}
\subsubsection{EVA3D}

EVA3D \cite{hong2022eva3d} is an unconditional 3D human generative model learned only from 2D image collections. At the core of EVA3D is a compositional NeRF representation, which divides the human body into local parts and each part is represented by an individual network modeling the corresponding local volume. With the compositional human NeRF representation, a 3D human GAN framework is constructed where the generator is defined as follows:
\begin{equation}
G(z, \beta, \theta, \text{cam}) = R(F(\Phi_G(z), \beta, \theta, \text{cam}))
\end{equation}
where $z$ represents noise sampled from a normal distribution, $\text{cam}$ denotes the camera pose, and $(\beta, \theta)$ are SMPL parameters controlling body shapes. These parameters are sampled from estimated distributions that are derived from 2D image collections.

The rendering algorithm $R(F_{\Phi_G}(z), \beta, \theta, \textit{cam})$ is responsible for generating the final output. Each sub-network within the compositional NeRF representation \(F_{\Phi}\) is composed of stacked MLPs with SIREN activation \cite{sitzmann2020implicit}. To generate fake samples, noise \(z \sim p_z\), SMPL parameters ($\beta, \theta$) , together with camera poses \textit{cam} are sampled. The generator then produces fake samples \(I_f = G(z, \beta, \theta, \textit{cam}; \Phi_G)\), which, along with real samples \(I_r \sim p_{\text{real}}\), are sent to discriminator \(D(I; \Phi_D)\) for adversarial training. 

\subsubsection{Score Distillation Sampling}
Score Distillation Sampling (SDS), first proposed in DreamFusion \cite{poole2022dreamfusion}, distills the prior knowledge from a pretrained diffusion model $\bm{\epsilon}_{\mathrm{\phi}}$ into a differentiable 3D representation $\mathit{\theta}$ given an input text prompt $\textit{y}$. SDS optimizes the 3D representation such that its multi-view renderings resemble high-quality samples from a frozen diffusion model. Specifically, given a rendered image $\mathit{I} = \mathit{g}(\theta)$ from a differentiable 3D model \textit{g}, random noise $\bm{\epsilon}$ is added to obtain a noisy image. The SDS loss then computes the gradient of $\mathit{\theta}$ by minimizing the difference between the predicted noise
$\bm{\epsilon}_{\phi} (x_t; y, t)$ and the added noise $\bm{\epsilon}$ , which can be formulated by:
\begin{equation}
\nabla_{\theta} L_{SDS}(\phi, x_0) = \mathbb{E}_{t, \epsilon} \left[ w(t) \left( \boldsymbol{\epsilon}(z_t; y_t) - \boldsymbol{\epsilon} \right) \frac{\partial x}{\partial \theta} \right]
\end{equation}
where $\mathit{z_t}$ denotes the noisy image at noise level \textit{t}, \textit{w(t)} is a
weighting function that depends on the noise level \textit{t} and the text prompt \textit{y}. In practice, using a strong classifier-free guidance (CFG \cite{ho2022classifier}) setting is crucial for diffusion models to create high-quality 3D samples, although it comes at the expense of generation diversity.

\subsection{DivAvatar}

Our rationale for adopting the unconditional human generative model (EVA3D) stems from the inherent capabilities of adversarial training: when coupled with stochastic noise inputs, GANs have the potential to yield outputs of considerable variation and distinction in inference time. EVA3D's inherent priors for human figure geometry and appearance make it an ideal starting point for further text-based conditioning. During training and inference of the original unconditional EVA3D, noise is randomly sampled from normal distribution as input, allowing unconditional diverse 3D human generation. 
 
To condition the pretrained generative model (i.e. EVA3D) on textual description, we integrate SDS loss into the fine-tuning process. However, our empirical observations indicate that the naive incorporation of SDS loss, while adept at generating 3D human representations aligned with textual descriptions, makes the generative model lose its ability to generate diverse samples (\cref{sub:ablation}). Given different noise samples in inference, the model always generates avatars with similar appearances with subtle differences. To address this issue, we design an alternative noise sampling technique that effectively resurrects the finetuned generative model's capability for diverse avatar generation from a single text prompt. The second challenge arises in the form of generation quality in terms of compromise of texture fidelity and unnatural dents in geometry. To address these issues, we introduce a comprehensive semantic aware zoom to bolster textual alignment in textures, and a cutting-edge feature-based depth loss to refine geometry respectively.
The overall framework of DivAvatar is shown in \cref{fig:pipeline}.

\subsubsection{Strategic Noise Sampling} 

The limitation in diversity in the finetuned generative model arises from the nature of SDS, which tends to exhibit a mode-seeking behavior combined with the high guidance weights \cite{wang2023prolificdreamer,poole2022dreamfusion}. In essence, SDS gravitates toward generating representations that share a common appearance, irrespective of the variations present in the input noise. When SDS is used with generative model, the model will update in a way that settles into a single mode or a few modes of the data distribution, ignoring the richness of the entire distribution. Therefore, even though we sample random noises as input to the human generative model (i.e EVA3D), there is little diversity in the generated samples.

Different from the traditional noise sampling in GAN training where a new noise is sampled in each iteration, we propose to alternate between sampling a random noise and a fixed noise throughout the finetuning process with the probabilities of \textit{p} and \textit{(1-p)} respectively. We primarily set \textit{p} to be 0.1 to ensure that the input noise remains constant most of the time. During inference, random noise is sampled from normal distribution and our finetuned generative model is able to generate avatars with varied appearances, each of which corresponds accurately to the textual prompt. This noise sampling method, while seemingly counter-intuitive, proves to be simple and effective, allowing our finetuned generative model to generate diverse outputs even in the face of the strong mode-seeking property of SDS. 

By predominantly holding the noise constant, the generator consistently receives the same source of randomness during training. This acts as a form of regularizer - the generator is deterred from seeking shortcuts using fluctuating noise patterns to mislead the discriminator. Instead of navigating the challenges of a constantly changing noise vector, the generator has a stable reference, enabling it to concentrate its efforts on learning transformations to produce varied samples. In scenarios with variable noise, the generator might quickly drift towards specific modes in the data to minimize the SDS loss, leading to mode collapse.  While the fixed noise forms the bedrock of our method, we intersperse it with random noise sampling in around 10\% of the iterations. Our training strategy allows one to have the option to generate both diverse and uniform avatar appearances by manipulating the value of \textit{p}, more details to be discussed in Section \ref{sub:ablation}.

\subsubsection{Semantic Zoom}
We observed that SDS loss sometimes generates avatars that do not fully match complex text inputs, such as those with involving multiple colors. To address this issue, we introduce a semantic-aware zoom on the compositional human NeRF which acts as an `attention map' guiding the SDS loss. While existing works incorporate a similar zoom mechanism \cite{kolotouros2023dreamhuman,zhang2023avatarverse}, they mainly utilize it to improve on blurry texture. In addition, our semantic zoom consists of a more comprehensive set of angles while the existing works often neglect the backviews. In the original EVA3D implementation, each body part is mapped to a unique SIREN network. This clear delineation simplifies the process of identifying individual body parts, rendering our semantic zoom approach for each part an intuitive solution. In total, we define 6 semantic regions: upper body and lower body with their respective back views, left hand and right hand. Based on the regions of interest, we also modify the text prompt accordingly so that the correct information can be encoded into the text. By including the back view as part of the semantic regions, we effectively eliminate the Janus problem present in other avatar works. Our semantic zoom effectively handles long or complex prompts, focusing the diffusion model on specific areas accurately.

\subsubsection{Loss Functions}

\textbf{Feature-Based Depth Loss.} Depth loss is exploited to avoid unnatural caved-in structures in the 3D geometry. Even though the existing loss functions in EVA3D is sufficient for unconditional 3D human generation, the SDS loss gives rise to curved-in geometry in our fine-tuning process. This problem becomes noticeable when watermarked textures (artifact due to EVA3D's SIREN generator \cite{hong2022eva3d}) mistakenly influence the geometry, creating unintended dents in the mesh generated. We alleviate this issue by designing a novel masked depth regularization in feature space of a VGG network \cite{simonyan2015deep}. Specifically, we adopt an off-the-shelf depth estimator \cite{Ranftl2021,Ranftl2022} to predict the depth map of rendered images as the `ground truth'. We then minimise the \(\mathcal{L}_1 \) loss of the feature maps of the masked `ground truth' depth and masked actual generated depth, which is formulated as:
\begin{equation}
\mathcal{L}_{\text{depth}} = \frac{1}{N} \sum_{i=1}^{N} | F(\text{depth}_{\text{predicted}})_i - F(\text{depth}_{\text{gt}})_i |
\end{equation}
where \( F(\cdot) \) represents the VGG feature extraction function, \( \text{depth}_{\text{predicted}} \) is the depth output by the generative model, \( \text{depth}_{\text{gt}} \) is the `ground truth' depth from the estimator, and \( N \) is the number of pixels in the masked area. This feature-based depth loss effectively eliminates the unintended dents due to watermark artifacts and smooths the generated surface.

\noindent
\textbf{Total Loss.} Beyond the aforementioned feature-based depth loss and the SDS loss, we utilise the existing loss functions present in EVA3D in the finetuning.
The overall loss function is defined as:
\begin{equation}
\mathcal{L}_{total} = \mathcal{L}_{eva3d} + \lambda_{sds} \mathcal{L}_{sds} + \lambda_{depth} \mathcal{L}_{depth}
\end{equation}
where \(\mathcal{L}_{eva3d}\) denote EVA3D loss, and \( \lambda_* \) are loss weights defined empirically.

\subsubsection{Mesh Optimization}

While our finetuned generative model generates humans with plausible geometry and appearance, its texture exhibits over-smoothness, saturated colors, and is not fully photorealistic \cite{poole2022dreamfusion} due to the high CFG \cite{ho2022classifier} values we employ to guarantee generation quality. Inspired by Magic3D \cite{lin2023magic3d} and Magic123 \cite{qian2023magic123}, we further export the mesh obtained from our finetuned model to the texture-decomposed SDF-Mesh hybrid representation (DMTet \cite{shen2021deep}). This allows us to learn textured 3D meshes of the generated avatars from rendered images, utilizing the same diffusion prior, SDS loss. Additionally, we refine the rendered images from the finetuned generative model using an img2img pipeline \cite{img2img} to add photorealistic details. These refined images are used as image conditions to ensure that the diverse appearances will not be lost in the process of DMTet finetuning. We achieve this by minimising the MSE loss between rendered textured mesh and the input images. 

\section{Experiments}
\label{sec:expt}

\begin{figure*}[hbt!]
  \centering
   \includegraphics[width=0.9\linewidth]{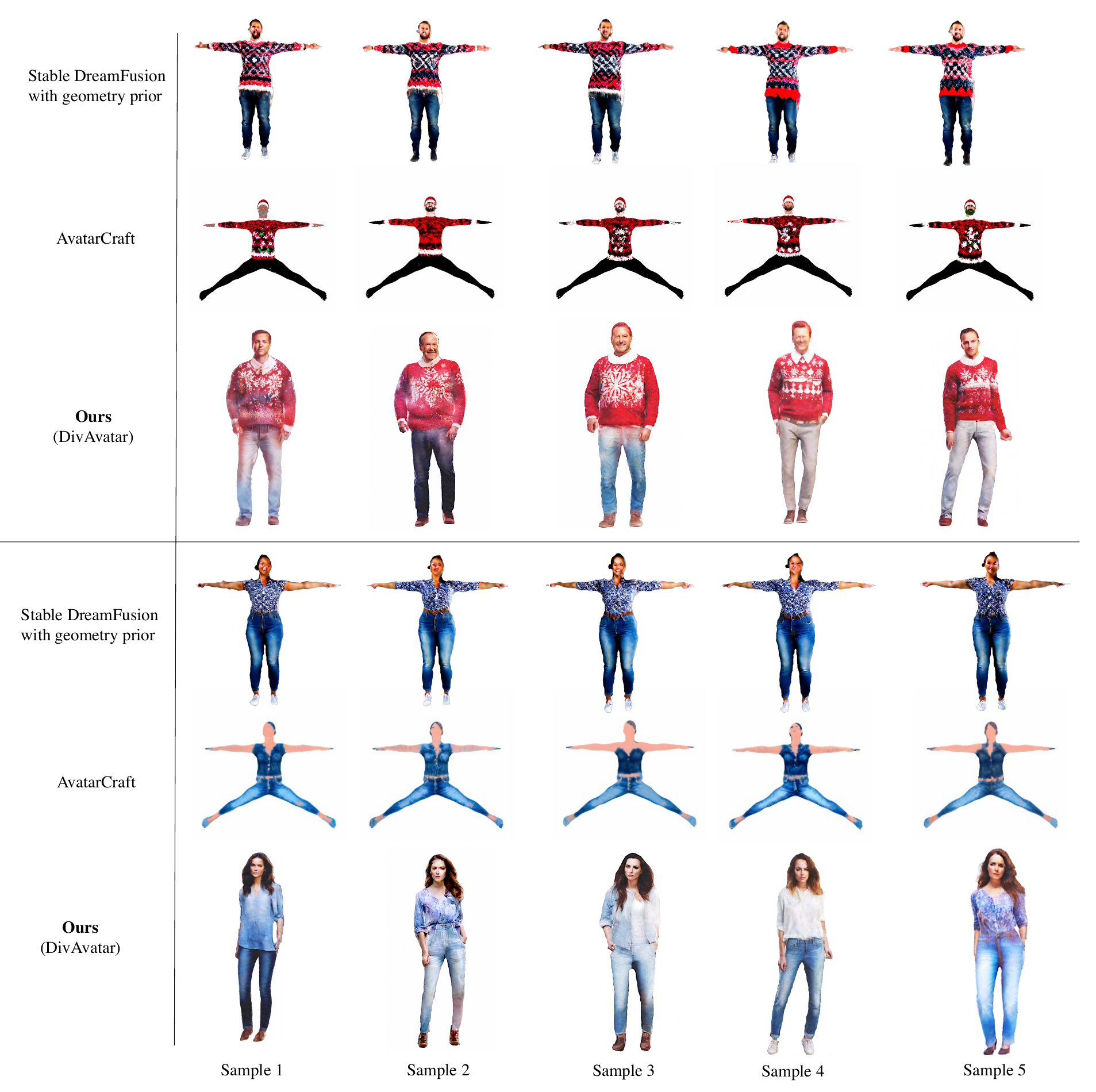}
   \caption{We demonstrate qualitative results for two different prompts with two existing methods. For each prompt, we obtain five different samples. Our method generates avatars that are more diverse in appearances and can be readily posed to natural poses. Input text prompt (top): \textit{A man wearing Christmas sweater. }Input text prompt (bottom): \textit{A woman wearing denim.}}
   \label{fig:xmas}
\end{figure*}

\begin{figure*}[h!]
  \centering
   \includegraphics[width=0.9\linewidth]{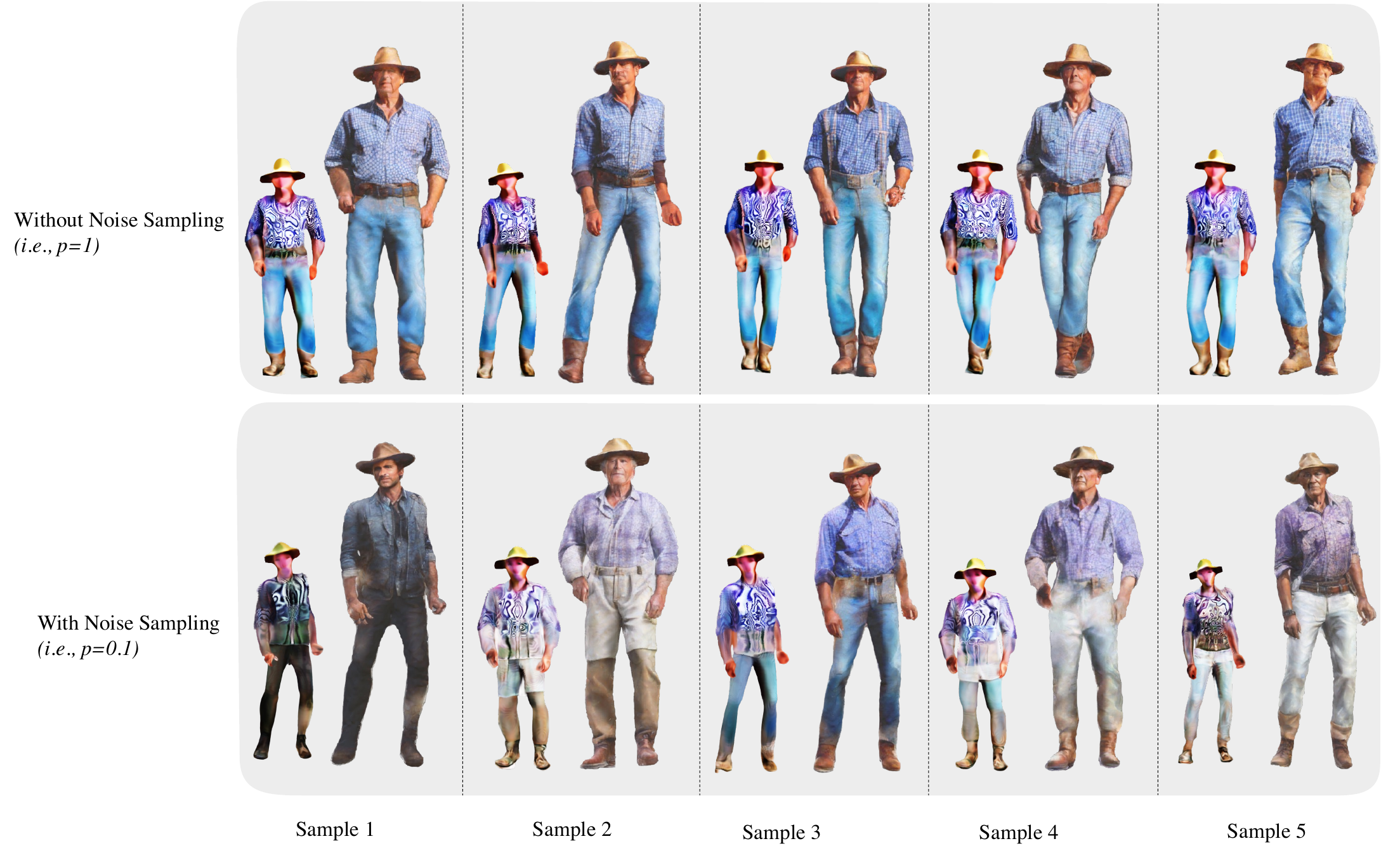}
   \caption{Importance of our noise sampling. Input text prompt: \textit{A farmer.} We show results across 5 inference samples. Without our noise sampling \textit{(i.e, p=1)}, the model generates avatars with little variations (top row). When we fix the noise sampling in majority of the time \textit{(i.e, p=0.1)}, the generated avatars are much more varied. For each sample, we show the coarse avatar from finetuned EVA3D (left) and the refined avatar after mesh optimization (right). Even though mesh optimization adds photorealistic texture details, the diversity stems from our noise sampling method. }
   \label{fig:sample}
\end{figure*}

\begin{figure}[]
  \centering
   \includegraphics[width=0.8
\linewidth]{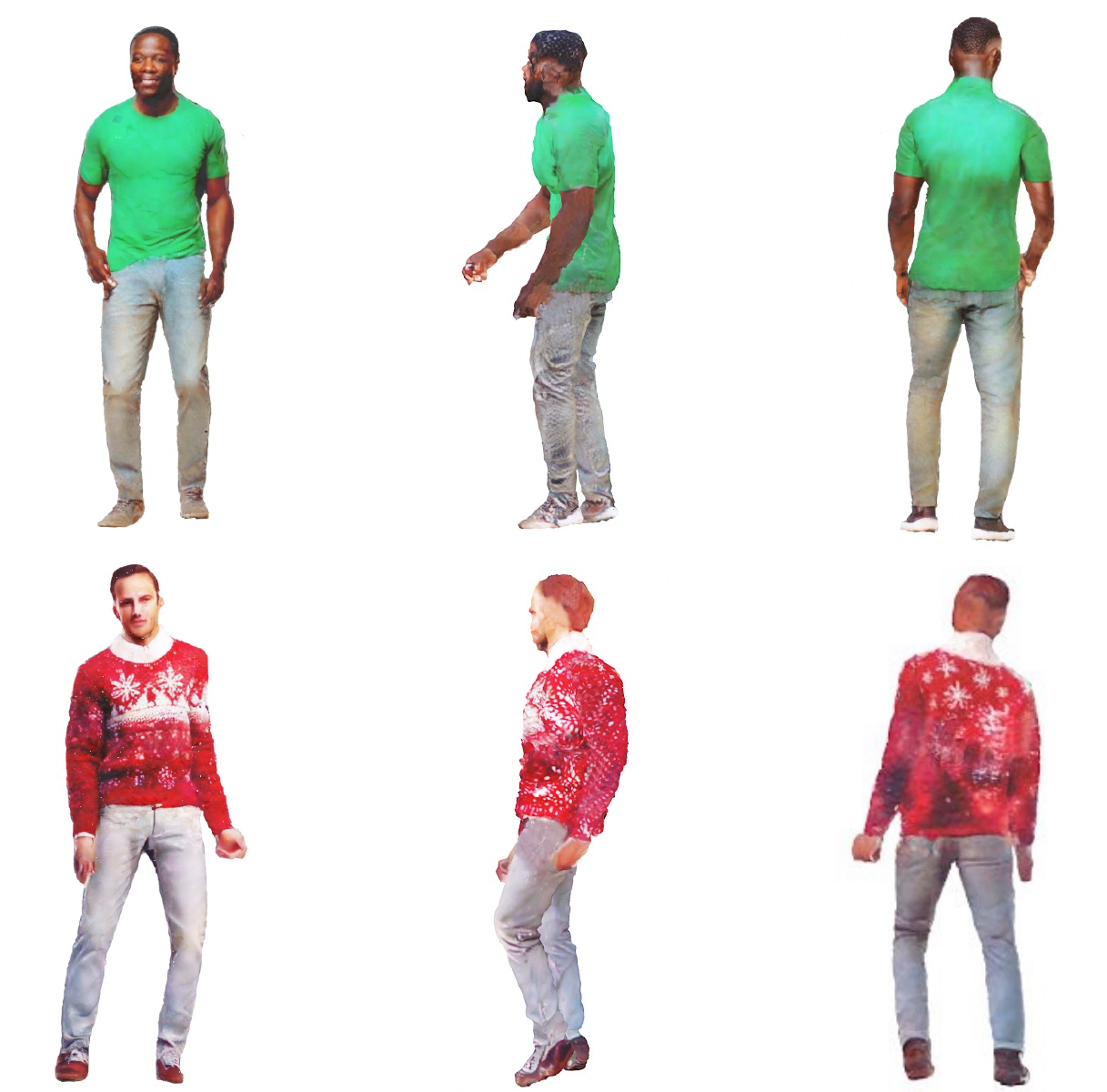}

   \caption{Multiview results. Input text prompt of top row: \textit{A Black man wearing green tshirt.} Input text prompt of bottom row: \textit{A man wearing Christmas sweater.}}

   \label{fig:multi}
\end{figure}

\subsection{Implementation Details}

We use the publicly available release version v2.0 of Stable Diffusion \cite{rombach2022high} as the diffusion prior for SDS loss, with the CFG value of 100. We utilise the EVA3D model trained with DeepFashion dataset \cite{liu2016deepfashion} over 420000 iterations. We then finetune this generative model for 5000 iterations and setting \textit{p} (probability of sampling random noise) to be 0.1, \(\lambda_\text{sds} \) and \( \lambda_\text{depth} \) to be 1. We conduct experiments on one Tesla V100 GPU.

\begin{figure*}[h!]
  \centering
   \includegraphics[width=1\linewidth]{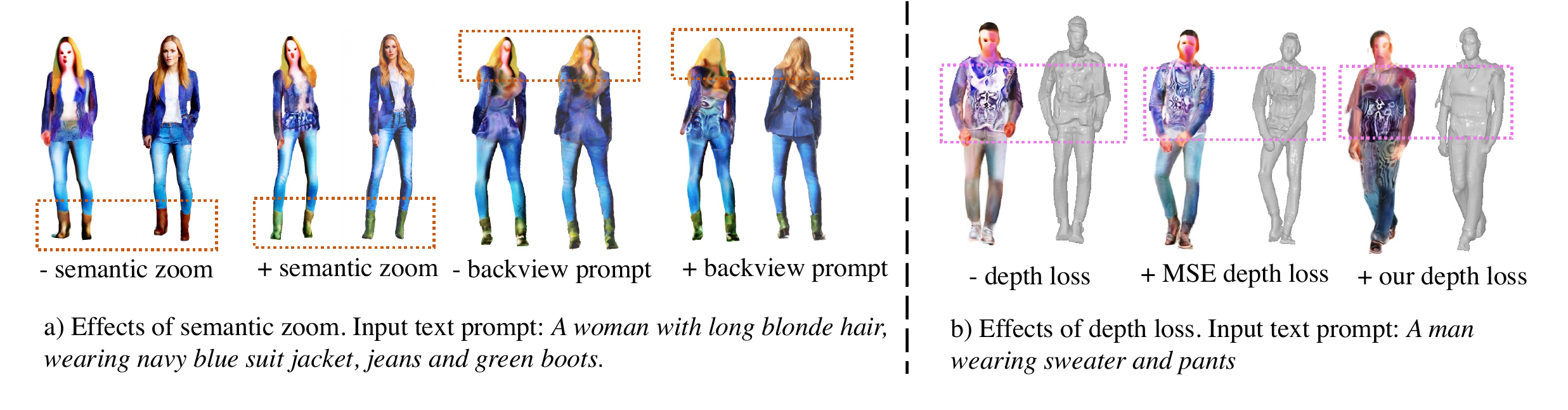}
   \caption{Ablation study on semantic zoom (a) and feature-based depth loss (b). In (a), we show the rendered image from finetuned generative model (i.e EVA3D) (left) and after mesh optimization (right) in each pair. Without the semantic zoom, the finetuned generative model is unable to capture the right information on color of boots. Addition of backview prompt improves the backview head quality . In (b), we show the rendered image (left) and untextured mesh (right) from finetuned generative model (EVA3D). Our feature-based depth loss effectively smooths out unnatural dents in the geometry caused by watermarks artifacts in the texture.}
   \label{fig:semantic}
\end{figure*}

\subsection{Qualitative Results}

In \cref{fig:multi}, we show multi-view images from our work, which exhibits high quality and multi-view consistent appearances that align well to text prompt.

We show a qualitative comparison of our method with existing works in \cref{fig:xmas}. As most avatar works did not release their code, we are unable to conduct repeated experiments to compare their diversity. We therefore conduct the qualitative comparison on Stable Dreamfusion (ICLR 2022) and AvatarCraft (ICCV 2023) whose codes are available. We use the publicly available code of Stable Dreamfusion \cite{stable-dreamfusion}, which replaces the original Imagen model \cite{saharia2022photorealistic} in Dreamfusion \cite{poole2022dreamfusion} with the Stable Diffusion model \cite{rombach2022high}. As the baseline Stable Dreamfusion does not have any geometry prior, it generates unrecognizable boundaries that does not have the shape of a human. Hence, we incorporate a zero-pose human body model as a geometry prior for the experiments. Additionally, we finetune it with DMTet (as in ours) for better performance and fair comparison. For each text prompt, we trained the Stable Dreamfusion (finetuned with DMTet) and AvatarCraft model five times to obtain five sets of results. On the other hand, we train DivAvatar once and obtain five sets of results by randomly sampling five noises in normal distribution in inference time, followed by mesh optimization of each avatar sample. Our result demonstrates a significantly wider array of appearances that closely match the input text prompt, in contrast to AvatarCraft and Stable Dreamfusion that produce results with only minor variations. Furthermore, our method is able to generate posed avatars in any sampled poses, whereas AvatarCraft and Stable Dreamfusion generate avatars in their input poses and need to be further articulated for other poses. We demonstrate more comparisons in the Supplementary Materials.

\subsection{Ablation Study}
\label{sub:ablation}
We conduct ablation study on the aforementioned designs in the following section.

\textbf{Strategic Noise Sampling.} We highlight the critical role of noise sampling in producing diverse outputs in \cref{fig:sample}. Utilizing the SDS loss with random noise sampling during finetuning leads to generated avatars with only minor variations, regardless of the noise introduced at the time of inference. This is evident in the farmer avatars (top row), which display strikingly similar looks across five separate inference attempts. However, with the introduction of our crafted noise sampling strategy, the generator's ability to produce a wide range of distinct avatars is revived. Consequently, this results in a collection of farmer avatars, each with a unique and recognizable appearance (bottom row). By adjusting \textit{p}, the probability of sampling a new noise in each iteration, one can have higher control on the level of diversity in the avatar generation. It is important to note that the diversity stems from our noise sampling approach (left image of each sample), instead of the seemingly contributing factor of mesh optimization (right image of each sample). This is elaborated in the following part of ablation study on Mesh Optimization.

\textbf{Semantic Aware Zoom.} In \cref{fig:semantic}, we show the importance of our semantic aware zoom. This zoom mechanism improves the textual fidelity, particularly when the text prompt is long. Without the semantic aware zoom, the generator may miss out on parts of the text prompt and generates appearances that is not well aligned with the prompt text. The final appearance will still be inaccurate even after the subsequent mesh optimization.

\textbf{VGG Feature-Based Depth Loss.} We present the rendered untextured mesh for geometry comparison in \cref{fig:semantic} to demonstrate the effectiveness of our feature-based depth loss. Without our feature-space depth loss, the generated mesh erroneously exhibits indentations mirroring the water-print patterns found on the clothing texture (left side). Such patterns should only exist in the texture instead of altering the mesh's geometry. Comparing with the MSE depth loss, Our feature-based depth loss eliminates these inappropriate indentations better, ensuring a smoother mesh surface regardless of the texture's water-print artifacts. 

\textbf{Mesh Optimization.}
We use mesh optimization to enhance the textures and photorealism of avatars, as shown in \cref{fig:sample}, where the left and right images of each sample display the pre- and post-optimization effects, respectively. However, it is important to note that the mesh optimization, while contributing to photorealistic details, does not by itself contribute to diverse appearances. This is illustrated in the top row of \cref{fig:sample}: avatars with only minor differences in their initial, coarse form (left side image of each sample) continue to exhibit similar appearances after mesh optimization (right side image of each sample). True diversity in textures emerges only when the initial avatars are generated using our noise sampling method, resulting in noticeable variations.


 

\section{Conclusion}

In conclusion, DivAvatar, our novel framework, marks a significant advancement in text-to-avatar generation by addressing the diversity challenge prevalent in current avatar methods. By fine-tuning a human generative model (EVA3D) and employing unique noise sampling during training, DivAvatar excels in generating a broad spectrum of distinct avatars from a single text prompt. Its key features - innovative noise sampling for training, a semantic-aware zoom mechanism, and feature-based depth loss, are instrumental in achieving both diversity in appearances and enhanced visual quality. Our results demonstrate DivAvatar's exceptional capability to empower creatives with varied and detailed 3D avatars, aligned closely with textual descriptions.

\noindent
\textbf{Limitations and future work.} Even though we utilise a GAN model for diverse appearances in inference times, the output texture lacks photorealistic details, necessitating additional mesh optimization for each sample. Enhancing texture quality for direct high-quality avatar generation without further refinement remains for future work. Additionally, our model shows limited diversity in specific uniforms like nurses or firefighters, likely due to reliance on appearance priors from the EVA3D's training on the DeepFashion dataset. If prompt subject is not in the training dataset, the main source of appearance will be dependent on the SDS loss which tends to converge. Furthermore, we encounter ongoing challenges inherent to EVA3D, such as watermark artifacts and inability to generate loose clothes.
 
{
    \small
    \bibliographystyle{ieeenat_fullname}
    \bibliography{main}

\begin{thebibliography}{35}
\providecommand{\natexlab}[1]{#1}
\providecommand{\url}[1]{\texttt{#1}}
\expandafter\ifx\csname urlstyle\endcsname\relax
  \providecommand{\doi}[1]{doi: #1}\else
  \providecommand{\doi}{doi: \begingroup \urlstyle{rm}\Url}\fi

\bibitem[img()]{img2img}
Stable diffusion xl 1.0 image to image pipeline cpu.
\newblock \url{https://huggingface.co/spaces/Manjushri/SDXL-1.0-Img2Img-CPU}.
\newblock Accessed: 2023-11-15.

\bibitem[Brown et~al.(2020)Brown, Mann, Ryder, Subbiah, Kaplan, Dhariwal, Neelakantan, Shyam, Sastry, Askell, et~al.]{brown2020language}
Tom Brown, Benjamin Mann, Nick Ryder, Melanie Subbiah, Jared~D Kaplan, Prafulla Dhariwal, Arvind Neelakantan, Pranav Shyam, Girish Sastry, Amanda Askell, et~al.
\newblock Language models are few-shot learners.
\newblock \emph{Advances in neural information processing systems}, 33:\penalty0 1877--1901, 2020.

\bibitem[Cao et~al.(2023)Cao, Cao, Han, Shan, and Wong]{cao2023dreamavatar}
Yukang Cao, Yan-Pei Cao, Kai Han, Ying Shan, and Kwan-Yee~K Wong.
\newblock Dreamavatar: Text-and-shape guided 3d human avatar generation via diffusion models.
\newblock \emph{arXiv preprint arXiv:2304.00916}, 2023.

\bibitem[Ho and Salimans(2022)]{ho2022classifier}
Jonathan Ho and Tim Salimans.
\newblock Classifier-free diffusion guidance.
\newblock \emph{arXiv preprint arXiv:2207.12598}, 2022.

\bibitem[Ho et~al.(2020)Ho, Jain, and Abbeel]{ho2020denoising}
Jonathan Ho, Ajay Jain, and Pieter Abbeel.
\newblock Denoising diffusion probabilistic models.
\newblock \emph{Advances in neural information processing systems}, 33:\penalty0 6840--6851, 2020.

\bibitem[Hong et~al.(2022{\natexlab{a}})Hong, Chen, Lan, Pan, and Liu]{hong2022eva3d}
Fangzhou Hong, Zhaoxi Chen, Yushi Lan, Liang Pan, and Ziwei Liu.
\newblock Eva3d: Compositional 3d human generation from 2d image collections.
\newblock \emph{arXiv preprint arXiv:2210.04888}, 2022{\natexlab{a}}.

\bibitem[Hong et~al.(2022{\natexlab{b}})Hong, Zhang, Pan, Cai, Yang, and Liu]{hong2022avatarclip}
Fangzhou Hong, Mingyuan Zhang, Liang Pan, Zhongang Cai, Lei Yang, and Ziwei Liu.
\newblock Avatarclip: Zero-shot text-driven generation and animation of 3d avatars.
\newblock \emph{arXiv preprint arXiv:2205.08535}, 2022{\natexlab{b}}.

\bibitem[Huang et~al.(2023)Huang, Wang, Zeng, Cao, Qi, Shi, Zha, and Zhang]{huang2023dreamwaltz}
Yukun Huang, Jianan Wang, Ailing Zeng, He Cao, Xianbiao Qi, Yukai Shi, Zheng-Jun Zha, and Lei Zhang.
\newblock Dreamwaltz: Make a scene with complex 3d animatable avatars.
\newblock \emph{arXiv preprint arXiv:2305.12529}, 2023.

\bibitem[Jain et~al.(2022)Jain, Mildenhall, Barron, Abbeel, and Poole]{jain2022zero}
Ajay Jain, Ben Mildenhall, Jonathan~T Barron, Pieter Abbeel, and Ben Poole.
\newblock Zero-shot text-guided object generation with dream fields.
\newblock In \emph{Proceedings of the IEEE/CVF Conference on Computer Vision and Pattern Recognition}, pages 867--876, 2022.

\bibitem[Jiang et~al.(2023)Jiang, Wang, Zhang, Chai, He, Chen, and Liao]{jiang2023avatarcraft}
Ruixiang Jiang, Can Wang, Jingbo Zhang, Menglei Chai, Mingming He, Dongdong Chen, and Jing Liao.
\newblock Avatarcraft: Transforming text into neural human avatars with parameterized shape and pose control.
\newblock \emph{arXiv preprint arXiv:2303.17606}, 2023.

\bibitem[Kolotouros et~al.(2023)Kolotouros, Alldieck, Zanfir, Bazavan, Fieraru, and Sminchisescu]{kolotouros2023dreamhuman}
Nikos Kolotouros, Thiemo Alldieck, Andrei Zanfir, Eduard~Gabriel Bazavan, Mihai Fieraru, and Cristian Sminchisescu.
\newblock Dreamhuman: Animatable 3d avatars from text.
\newblock \emph{arXiv preprint arXiv:2306.09329}, 2023.

\bibitem[Lin et~al.(2023)Lin, Gao, Tang, Takikawa, Zeng, Huang, Kreis, Fidler, Liu, and Lin]{lin2023magic3d}
Chen-Hsuan Lin, Jun Gao, Luming Tang, Towaki Takikawa, Xiaohui Zeng, Xun Huang, Karsten Kreis, Sanja Fidler, Ming-Yu Liu, and Tsung-Yi Lin.
\newblock Magic3d: High-resolution text-to-3d content creation.
\newblock In \emph{Proceedings of the IEEE/CVF Conference on Computer Vision and Pattern Recognition}, pages 300--309, 2023.

\bibitem[Liu et~al.(2016)Liu, Luo, Qiu, Wang, and Tang]{liu2016deepfashion}
Ziwei Liu, Ping Luo, Shi Qiu, Xiaogang Wang, and Xiaoou Tang.
\newblock Deepfashion: Powering robust clothes recognition and retrieval with rich annotations.
\newblock In \emph{Proceedings of the IEEE conference on computer vision and pattern recognition}, pages 1096--1104, 2016.

\bibitem[Loper et~al.(2023)Loper, Mahmood, Romero, Pons-Moll, and Black]{loper2023smpl}
Matthew Loper, Naureen Mahmood, Javier Romero, Gerard Pons-Moll, and Michael~J Black.
\newblock Smpl: A skinned multi-person linear model.
\newblock In \emph{Seminal Graphics Papers: Pushing the Boundaries, Volume 2}, pages 851--866. 2023.

\bibitem[master Chef and master Kratos()]{masteravatarstudio}
Karate master Chef and Karate master Kratos.
\newblock Avatarstudio: High-fidelity and animatable 3d avatar creation from text.

\bibitem[Metzer et~al.(2023)Metzer, Richardson, Patashnik, Giryes, and Cohen-Or]{metzer2023latent}
Gal Metzer, Elad Richardson, Or Patashnik, Raja Giryes, and Daniel Cohen-Or.
\newblock Latent-nerf for shape-guided generation of 3d shapes and textures.
\newblock In \emph{Proceedings of the IEEE/CVF Conference on Computer Vision and Pattern Recognition}, pages 12663--12673, 2023.

\bibitem[Mildenhall et~al.(2021)Mildenhall, Srinivasan, Tancik, Barron, Ramamoorthi, and Ng]{mildenhall2021nerf}
Ben Mildenhall, Pratul~P Srinivasan, Matthew Tancik, Jonathan~T Barron, Ravi Ramamoorthi, and Ren Ng.
\newblock Nerf: Representing scenes as neural radiance fields for view synthesis.
\newblock \emph{Communications of the ACM}, 65\penalty0 (1):\penalty0 99--106, 2021.

\bibitem[Mohammad~Khalid et~al.(2022)Mohammad~Khalid, Xie, Belilovsky, and Popa]{mohammad2022clip}
Nasir Mohammad~Khalid, Tianhao Xie, Eugene Belilovsky, and Tiberiu Popa.
\newblock Clip-mesh: Generating textured meshes from text using pretrained image-text models.
\newblock In \emph{SIGGRAPH Asia 2022 conference papers}, pages 1--8, 2022.

\bibitem[Poole et~al.(2022)Poole, Jain, Barron, and Mildenhall]{poole2022dreamfusion}
Ben Poole, Ajay Jain, Jonathan~T Barron, and Ben Mildenhall.
\newblock Dreamfusion: Text-to-3d using 2d diffusion.
\newblock \emph{arXiv preprint arXiv:2209.14988}, 2022.

\bibitem[Qian et~al.(2023)Qian, Mai, Hamdi, Ren, Siarohin, Li, Lee, Skorokhodov, Wonka, Tulyakov, et~al.]{qian2023magic123}
Guocheng Qian, Jinjie Mai, Abdullah Hamdi, Jian Ren, Aliaksandr Siarohin, Bing Li, Hsin-Ying Lee, Ivan Skorokhodov, Peter Wonka, Sergey Tulyakov, et~al.
\newblock Magic123: One image to high-quality 3d object generation using both 2d and 3d diffusion priors.
\newblock \emph{arXiv preprint arXiv:2306.17843}, 2023.

\bibitem[Radford et~al.(2021)Radford, Kim, Hallacy, Ramesh, Goh, Agarwal, Sastry, Askell, Mishkin, Clark, et~al.]{radford2021learning}
Alec Radford, Jong~Wook Kim, Chris Hallacy, Aditya Ramesh, Gabriel Goh, Sandhini Agarwal, Girish Sastry, Amanda Askell, Pamela Mishkin, Jack Clark, et~al.
\newblock Learning transferable visual models from natural language supervision.
\newblock In \emph{International conference on machine learning}, pages 8748--8763. PMLR, 2021.

\bibitem[Raffel et~al.(2020)Raffel, Shazeer, Roberts, Lee, Narang, Matena, Zhou, Li, and Liu]{raffel2020exploring}
Colin Raffel, Noam Shazeer, Adam Roberts, Katherine Lee, Sharan Narang, Michael Matena, Yanqi Zhou, Wei Li, and Peter~J Liu.
\newblock Exploring the limits of transfer learning with a unified text-to-text transformer.
\newblock \emph{The Journal of Machine Learning Research}, 21\penalty0 (1):\penalty0 5485--5551, 2020.

\bibitem[Ranftl et~al.(2021)Ranftl, Bochkovskiy, and Koltun]{Ranftl2021}
Ren\'{e} Ranftl, Alexey Bochkovskiy, and Vladlen Koltun.
\newblock Vision transformers for dense prediction.
\newblock \emph{ICCV}, 2021.

\bibitem[Ranftl et~al.(2022)Ranftl, Lasinger, Hafner, Schindler, and Koltun]{Ranftl2022}
Ren\'{e} Ranftl, Katrin Lasinger, David Hafner, Konrad Schindler, and Vladlen Koltun.
\newblock Towards robust monocular depth estimation: Mixing datasets for zero-shot cross-dataset transfer.
\newblock \emph{IEEE Transactions on Pattern Analysis and Machine Intelligence}, 44\penalty0 (3), 2022.

\bibitem[Rombach et~al.(2022)Rombach, Blattmann, Lorenz, Esser, and Ommer]{rombach2022high}
Robin Rombach, Andreas Blattmann, Dominik Lorenz, Patrick Esser, and Bj{\"o}rn Ommer.
\newblock High-resolution image synthesis with latent diffusion models.
\newblock In \emph{Proceedings of the IEEE/CVF conference on computer vision and pattern recognition}, pages 10684--10695, 2022.

\bibitem[Saharia et~al.(2022)Saharia, Chan, Saxena, Li, Whang, Denton, Ghasemipour, Gontijo~Lopes, Karagol~Ayan, Salimans, et~al.]{saharia2022photorealistic}
Chitwan Saharia, William Chan, Saurabh Saxena, Lala Li, Jay Whang, Emily~L Denton, Kamyar Ghasemipour, Raphael Gontijo~Lopes, Burcu Karagol~Ayan, Tim Salimans, et~al.
\newblock Photorealistic text-to-image diffusion models with deep language understanding.
\newblock \emph{Advances in Neural Information Processing Systems}, 35:\penalty0 36479--36494, 2022.

\bibitem[Sanghi et~al.(2022)Sanghi, Chu, Lambourne, Wang, Cheng, Fumero, and Malekshan]{sanghi2022clip}
Aditya Sanghi, Hang Chu, Joseph~G Lambourne, Ye Wang, Chin-Yi Cheng, Marco Fumero, and Kamal~Rahimi Malekshan.
\newblock Clip-forge: Towards zero-shot text-to-shape generation.
\newblock In \emph{Proceedings of the IEEE/CVF Conference on Computer Vision and Pattern Recognition}, pages 18603--18613, 2022.

\bibitem[Shen et~al.(2021)Shen, Gao, Yin, Liu, and Fidler]{shen2021deep}
Tianchang Shen, Jun Gao, Kangxue Yin, Ming-Yu Liu, and Sanja Fidler.
\newblock Deep marching tetrahedra: a hybrid representation for high-resolution 3d shape synthesis.
\newblock \emph{Advances in Neural Information Processing Systems}, 34:\penalty0 6087--6101, 2021.

\bibitem[Simonyan and Zisserman(2015)]{simonyan2015deep}
Karen Simonyan and Andrew Zisserman.
\newblock Very deep convolutional networks for large-scale image recognition, 2015.

\bibitem[Sitzmann et~al.(2020)Sitzmann, Martel, Bergman, Lindell, and Wetzstein]{sitzmann2020implicit}
Vincent Sitzmann, Julien Martel, Alexander Bergman, David Lindell, and Gordon Wetzstein.
\newblock Implicit neural representations with periodic activation functions.
\newblock \emph{Advances in neural information processing systems}, 33:\penalty0 7462--7473, 2020.

\bibitem[Song et~al.(2020)Song, Meng, and Ermon]{song2020denoising}
Jiaming Song, Chenlin Meng, and Stefano Ermon.
\newblock Denoising diffusion implicit models.
\newblock \emph{arXiv preprint arXiv:2010.02502}, 2020.

\bibitem[Tang(2022)]{stable-dreamfusion}
Jiaxiang Tang.
\newblock Stable-dreamfusion: Text-to-3d with stable-diffusion, 2022.
\newblock https://github.com/ashawkey/stable-dreamfusion.

\bibitem[Wang et~al.(2023)Wang, Lu, Wang, Bao, Li, Su, and Zhu]{wang2023prolificdreamer}
Zhengyi Wang, Cheng Lu, Yikai Wang, Fan Bao, Chongxuan Li, Hang Su, and Jun Zhu.
\newblock Prolificdreamer: High-fidelity and diverse text-to-3d generation with variational score distillation.
\newblock \emph{arXiv preprint arXiv:2305.16213}, 2023.

\bibitem[Zhang et~al.(2023{\natexlab{a}})Zhang, Chen, Yang, Qu, Wang, Chen, Long, Zhu, Du, and Zheng]{zhang2023avatarverse}
Huichao Zhang, Bowen Chen, Hao Yang, Liao Qu, Xu Wang, Li Chen, Chao Long, Feida Zhu, Kang Du, and Min Zheng.
\newblock Avatarverse: High-quality \& stable 3d avatar creation from text and pose.
\newblock \emph{arXiv preprint arXiv:2308.03610}, 2023{\natexlab{a}}.

\bibitem[Zhang et~al.(2023{\natexlab{b}})Zhang, Rao, and Agrawala]{zhang2023adding}
Lvmin Zhang, Anyi Rao, and Maneesh Agrawala.
\newblock Adding conditional control to text-to-image diffusion models.
\newblock In \emph{Proceedings of the IEEE/CVF International Conference on Computer Vision}, pages 3836--3847, 2023{\natexlab{b}}.

\end{thebibliography}
}



\end{document}